\documentclass[runningheads]{llncs}

\usepackage{eccv}

\usepackage{eccvabbrv}

\usepackage{graphicx}
\usepackage{booktabs}

\usepackage[accsupp]{axessibility}  
\usepackage{hyperref}
\usepackage{url}
\usepackage{graphicx}
\usepackage{booktabs}
\usepackage[table]{xcolor}
\usepackage{amsmath,amssymb,amsfonts}
\usepackage{multirow}
\usepackage{soul}
\usepackage{xcolor}
\sethlcolor{yellow!20}

\newcommand{\x}{\mathbf{x}}

\newcommand{\Lrec}{\mathcal{L}_\text{MSE}}
\newcommand{\Lmle}{\mathcal{L}_\text{MLE}}
\newcommand{\cupdot}{\mathbin{\mathaccent\cdot\cup}}

\usepackage{hyperref}

\usepackage{orcidlink}

\begin{document}

\title{MLE-UVAD: Minimal Latent Entropy Autoencoder for Fully Unsupervised Video Anomaly Detection}


\author{Yuang Geng$^{*}$ \and
Junkai Zhou$^{*}$ \and
Kang Yang$^{\dagger}$ \and
Pan He \and
Zhuoyang Zhou \and
Jose C. Principe \and
Joel Harley \and
Ivan Ruchkin}

\authorrunning{Y. Geng et al.}

\institute{University of Florida; Auburn University\\
$^{*}$Co-first authors.\quad $^{\dagger}$Corresponding author.}

\maketitle

\begin{abstract}

In this paper, we address the challenging problem of \textit{single-scene, fully unsupervised} video anomaly detection (VAD), where raw videos containing both normal and abnormal events are used directly for training and testing without any labels. This differs sharply from prior work that either requires extensive labeling (fully or weakly supervised) or depends on normal-only videos (one-class classification), which are vulnerable to distribution shifts and contamination. We propose an entropy-guided autoencoder that detects anomalies through reconstruction error by reconstructing normal frames well while making anomalies reconstruct poorly. The key idea is to combine the standard reconstruction loss with a novel Minimal Latent Entropy (MLE) loss in the autoencoder. Reconstruction loss alone maps normal and abnormal inputs to distinct latent clusters due to their inherent differences, but also risks reconstructing anomalies too well to detect. Therefore, MLE loss addresses this by minimizing the entropy of latent embeddings, encouraging them to concentrate around high-density regions. Since normal frames dominate the raw video, sparse anomalous embeddings are pulled into the normal cluster, so the decoder emphasizes normal patterns and produces poor reconstructions for anomalies. This dual-loss design produces a clear reconstruction gap that enables effective anomaly detection. Extensive experiments on two widely used benchmarks and a challenging self-collected driving dataset demonstrate that our method achieves robust and superior performance over baselines.

\end{abstract}

\section{Introduction}

Video anomaly detection (VAD) aims to identify abnormal events that deviate from regular patterns, such as violence, accidents, and other unexpected occurrences~\cite{VADliuSurvey,singlescenesurvay}.
Accurate detection is crucial for timely response in safety-critical applications, including surveillance~\cite{zhang2024trafficnight}, autonomous driving~\cite{geng2025unsupervised}, and unmanned aerial vehicles~\cite{2025privacyVADsurvey}.

Existing VAD approaches can be divided into three main categories~\cite{VADliuSurvey}: fully supervised~\cite{pang2020self}, weakly supervised~\cite{2024realweakly}, and unsupervised/one-class classification (OCC)~\cite{DAST_OCC}.
However, most existing methods rely on extensive human annotation.
Fully supervised approaches demand frame-level labels for every training video, which is prohibitively labor-intensive.
Weakly supervised methods ease this burden by using video-level labels—marking a video as abnormal if any frame contains an anomaly. Yet, such videos often include normal segments, making it difficult to disentangle abnormal frames from normal ones~\cite{tian2021weakly}.
A third line of work, unsupervised or one-class classification (OCC), trains exclusively on normal videos and requires no abnormal labels~\cite{zong2018deep,wang2021robust}. While this further reduces labeling costs, two major limitations remain in OCC:

\looseness=-1

\noindent
\textbf{Limitation 1: Sensitivity to Contamination.} 
  The effectiveness of OCC models depends on training exclusively with clean, anomaly-free data.
  Even a small fraction of anomalies in the training set can drastically degrade the detection performance~\cite{singlescenesurvay, 2019conceptYU}. 
  Therefore, compared to an unsupervised setting, OCC models require significant investment into a fully cleaned dataset, obtaining which is non-trivial and labor-intensive~\cite{CLAP_2024_CVPR}.

\noindent
\textbf{Limitation 2: High Cost for Labeling Normal Data.}
The OCC setting has no anomaly labels, but selecting ``normal'' videos is a subjective human task, the cost of which can vary across annotators~\cite{singlescenesurvay}.
Moreover, OCC methods assume that the distribution of normal videos remains consistent between training and testing~\cite{knowledgeinfuse_2025_CVPR}.
However, even minor shifts in camera characteristics, lighting conditions, or location can cause a distribution shift.
When training data distribution shifts, people need to relabel the normal-only videos from scratch.

To address these limitations and avoid any ambiguity of the supervision, we propose a \textit{fully unsupervised setting}, in which raw videos containing both normal and abnormal frames are used for training and testing without any labels, such as surveillance footage without manual annotation or privacy-sensitive labeling. 
Specifically, we focus on single-scene video anomaly detection where a fixed camera monitors a specific environment.
Most prior work overlooks the key distinction between single-scene and multi-scene settings~\cite{singlescenesurvay}.  
Multi-scene training uses videos from different environments and generalizes all scenes to what is normal vs. anomalous~\cite{multiscene-AD}. 
However, this contradicts the fact that anomalies are location-dependent~\cite{2024VADreview}: the same behavior may be normal in one region but abnormal in another (e.g., walking in a permitted area vs. a restricted zone).
In contrast, the single-scene setting focuses on the spatial or temporal regularities to learn location-dependent rules, yielding clearer decision boundaries and immediate practical use~\cite{multisceneUVA2024}.

Previous OCC methods often train autoencoders on normal videos with reconstruction loss, where higher reconstruction error flags out-of-distribution anomalies~\cite{DAST_OCC}.
Inspired by this, we address our fully unsupervised VAD problem with \textit{MLE-UVAD}, an entropy-guided autoencoder that augments the standard reconstruction loss with a minimal latent entropy (MLE) loss. 
With reconstruction loss alone in a fully unsupervised setting (e.g., unlabeled video), autoencoders often reconstruct both normal and anomalous frames accurately.
Therefore, no clear threshold on the reconstruction error can be set to detect anomalies. 
Next, the MLE was designed to maintain normal reconstructions while flagging anomalies by minimizing the latent entropy.

\begin{figure}[ht]
    \centering
    \includegraphics[width=\linewidth, clip, trim=0 210 0 0]{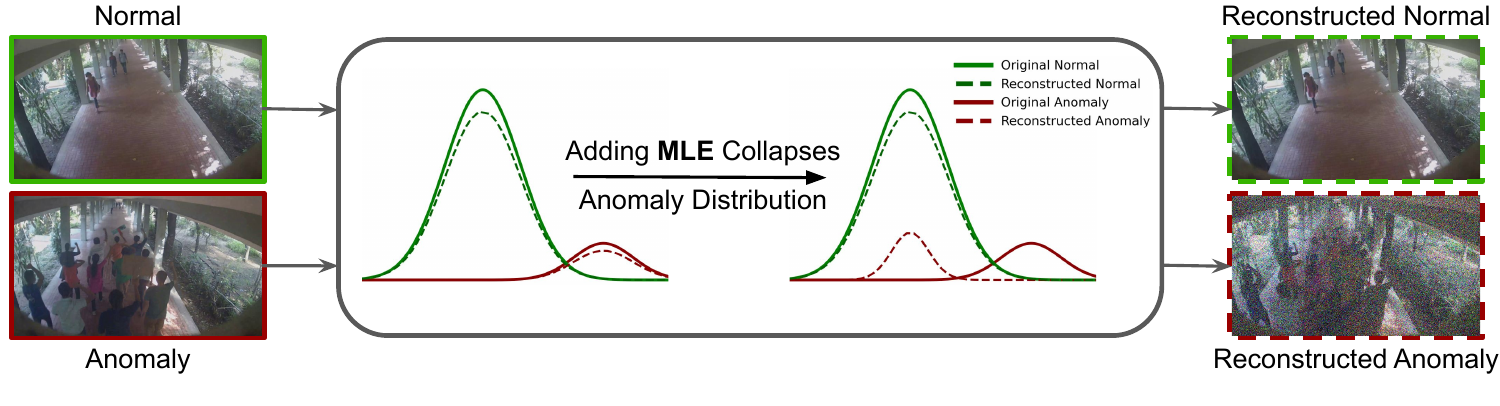}
    \caption{Collapse of the anomalous latent distribution. Green: normal frames. Red: abnormal frames. Solid lines: latent distributions of the original inputs. Dashed lines: latent distributions of the reconstructions. 
    The key idea is to pull sparse anomalous embeddings toward the dense normal embeddings, so normal frames remain well reconstructed while anomalies reconstruct poorly.
    }
\label{fig:mode_collaspe_final}
\end{figure}

Our key novelty is that we collapse anomalous latent representations toward the normal cluster with our proposed MLE loss, as shown in Fig.~\ref{fig:mode_collaspe_final}.
Mechanistically, reconstruction loss alone always maps normal and abnormal inputs to distinct latent distributions for accurate reconstruction.
The MLE loss regularizes these latent distributions by minimizing latent entropy.
Because normal videos dominate the unlabeled dataset, entropy minimization concentrates probability mass around the normal clusters and squeezes the sparse, shifted anomalous embeddings toward normal ones.
Therefore, anomalous frames are reconstructed as \textit{normal-looking} outputs and accordingly incur larger errors when compared to their original abnormal inputs.
Consequently, normal videos remain accurately reconstructed, whereas anomalous videos are poorly reconstructed.
This reconstruction gap provides a clear and practical threshold for anomaly detection.

\noindent
\textbf{Contributions.} The contributions of this paper are (1) A single-scene, fully unsupervised VAD setting that eliminates the need for human labeling; (2) An entropy-guided autoencoder with an MLE loss for fully unsupervised VAD; (3) An experimental validation on two public benchmarks and an autonomous racing dataset, consistently outperforming the baselines.

\section{Background and Related Work}

\paragraph{Weakly Supervised Video Anomaly Detection (WS-VAD).}            
WS-VAD assumes access to video-level labels: each training video is annotated as either normal or abnormal, but the time points of anomalies are unknown. 
This problem is typically formulated in a multiple-instance learning (MIL) framework~\cite{sultani2018real}, where a video is treated as a bag of snippets, and the model is trained to assign high anomaly scores to at least one snippet in abnormal bags. 
Subsequent works improved upon this paradigm by introducing contrastive objectives~\cite{tian2021weakly}, 
noise-robust label cleaning~\cite{zhong2019graph}, 
and discriminative representation learning~\cite{wan2021weakly}. 
However, WS-VAD still requires human annotation at the video level.
Moreover, abnormal-labeled videos often contain many normal snippets, resulting in significant label noise and difficulty in precisely localizing the anomalies. 


\paragraph{Unsupervised Video Anomaly Detection (UVAD).}
Most prior UVAD uses one-class classification: models train on normal-only data and flag deviations as anomalies~\cite{wang2021robust}.
They fall into three categories: reconstruction-based~\cite{reconstructionAD-oldest}, prediction-based~\cite{gan2predict2024}, and clustering-based~\cite{zong2018deep, 2024clustering}.
These methods are highly sensitive to the previous limitations: contaminated training data and distribution shift.

Fully unsupervised anomaly detection is quite sparse in the literature. Generative Cooperative Learning (GCL)~\cite{Zaheer_GCL} trains an autoencoder and a discriminator cooperatively on unlabeled videos, where pseudo-labels are iteratively refined to discourage anomaly reconstruction. 
While conceptually appealing, GCL can be unstable across scenarios due to a lack of clear pseudo-labeling rules, which accumulates errors to destabilize training.
Temporal Masked AutoEncoding (TMAE)~\cite{hu2022detecting} detects anomalies by training a Vision Transformer (ViT) to predict masked patches in spatiotemporal cubes of unlabeled video, where rare events yield higher prediction errors.
However, its reliance on ViT requires substantial computational overhead, hindering real-time deployment in our single-scene setting.
Both are compared with our method in the results.

\smallskip
\noindent
\textbf{Problem Formulation.}
Given the raw \textit{fully unsupervised video} directly from a fixed camera, our task is to detect whether an observation/frame $x_i$ is an anomaly. 
Given the video 
\begin{equation}
V = V^{\text{normal}} \cupdot V^{\text{abnormal}} = \{x_t\}_{t=1}^T, \qquad x_t \in \mathbb{R}^{C \times H \times W},
\end{equation}
where each observation $x_t$ is an image with $C$ channels and spatial resolution $H \times W$.
This problem requires calculating the anomaly score $S(x_t \mid V)$ given the whole video $V$ --- and then comparing it with a tunable threshold $\tau$. If the $S(x_t \mid V) > \tau$, then $x_t$ is flagged as an anomaly. 

\section{Methodology}
This section first overviews the MLE-UVAD pipeline and formalizes our proposed MLE loss.
Next, using t-SNE of the latent space, we show why it strengthens fully unsupervised learning.
Finally, we define a threshold on frame-wise reconstruction error for detecting anomalies.

\begin{figure}[htbp]
    \centering
    \includegraphics[width=\linewidth, trim=0 80 0 0]{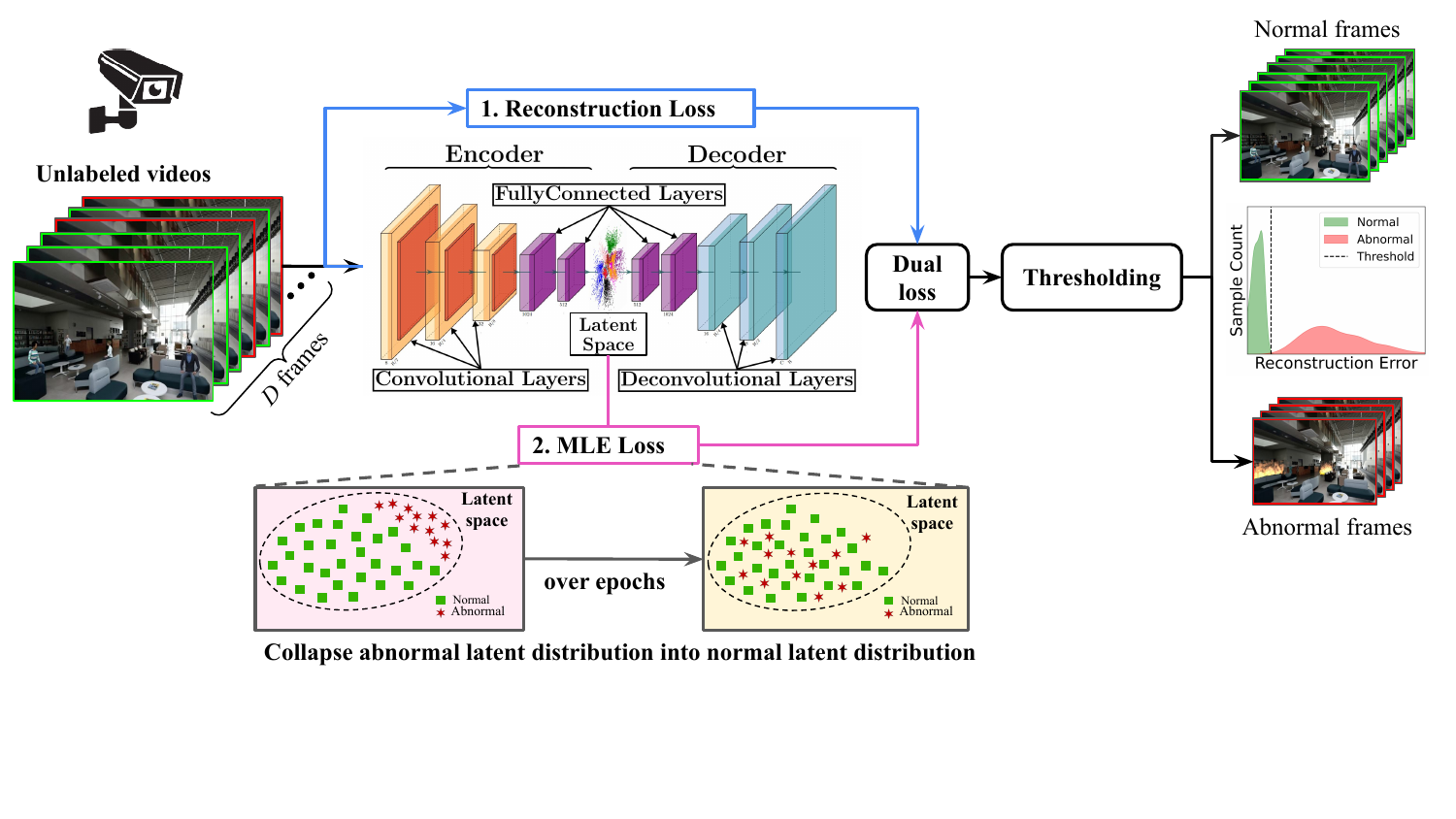}
    \caption{MLE-UVAD pipeline: unsupervised training and detection through reconstruction gap. Unlabeled videos are directly put into an autoencoder trained with a dual loss: (1) The MSE loss reconstructs all frames accurately.
    (2) The MLE collapses sparse abnormal embeddings toward the dense normal embedding, making anomaly reconstructions worse relative to normals. An anomaly is detected by setting a threshold on the reconstruction error.}
    \label{fig:overview_fig}
\end{figure}
\subsection{Overall MLE-UVAD Structure}
Our MLE-UVAD framework is designed to separate anomalies in the unlabelled video with a convolutional autoencoder (CAE), as shown in Fig.~\ref{fig:overview_fig}. 
This autoencoder serves as the reconstruction foundation~\cite{2023VAEcasual}: it encodes frames into latent
representations and decodes them back to the image space.
First, the reconstruction loss enables accurate video reconstruction, where normal and abnormal videos typically occupy different distribution regions.
Second, the MLE loss then reduces entropy in the latent space, collapsing abnormal embeddings into the dominant normal cluster. 
Together, these losses guarantee that normal frames are reconstructed well while abnormal frames are reconstructed poorly, creating a clear reconstruction gap. 
Thus, the dual-loss objective balances two effects: MSE preserves normal reconstructions, while MLE enforces distributional regularization that magnifies the gap between normal and abnormal frames.

After training the autoencoders, we measure the reconstruction quality via the Pearson Correlation Coefficient (PCC) for detection.
PCC emphasizes structural variation and is invariant to global brightness changes, making it more suitable than MSE (sensitive to scale) and cosine similarity (sensitive to shift).
The PCC is calculated between original frames $\mathbf{x_i}$ and its reconstruction $\mathbf{x'_i}$:
\begin{equation}
\text{pcc}_i = \frac{(\mathbf{x}'_i - \bar{x}_i')^{\intercal} (\mathbf{x}_i - \bar{x}_i)}{||\mathbf{x}_i' - \bar{x}'_i||_2 ||\mathbf{x}_i - \bar{x}_i||_2},
\end{equation}
where $\mathbf{x}_i$ and $\bar{x}_i$ are the $i$-th original frame in the video and its mean, respectively, while  $\mathbf{x}'_i$ and $\bar{x}'_i$ are the reconstructed frame and its mean. A low PCC (i.e., high reconstruction error) typically indicates the presence of an anomaly.

\subsection{Part 1: Reconstruction Loss $\Lrec$}

The first loss, reconstruction Loss $\Lrec$, forces the autoencoder to learn a compact representation from entire frames of video, and reconstruct frames from it.
The autoencoder with only $\Lrec$ can usually reconstruct both normal and abnormal frames well in the unlabeled dataset if the training is fully supervised and the model is sufficiently large. 

Formally, the autoencoder consists of an encoder $\mathbf{e}_{\theta}(\mathbf{x}_i) = \mathbf{z}_i$ and a decoder $\mathbf{d}_{\theta'}(\mathbf{z}_i) = \mathbf{x}'_i$, where $\mathbf{x}_i$ denotes an input, $\mathbf{z}_i$ denotes its latent representation, and $\mathbf{x}'_i$ denotes the its reconstruction. The model is trained to minimize the reconstruction loss over all the $N$ input samples, defined as:
\begin{equation}
\mathcal{L}_\text{MSE}(\theta, \theta') = \frac{1}{N} \sum_{i=1}^{N} \| \x_i - \x'_i \|_2
\label{eq:reconstruction_loss}
\end{equation}

\subsection{Part 2: Minimal Latent Entropy loss $\Lmle$}
In addition to $\Lrec$, we introduce the \emph{Minimal Latent Entropy} loss $\Lmle$, which explicitly minimizes the uncertainty (spread) of the autoencoder's latent embeddings.
Entropy is a fundamental concept in information theory that quantifies the uncertainty of a random variable~\cite{2024entropy}.
In here, we quantify this uncertainty of the latent variable $z$ via the Rényi entropy~\cite{renyi1961measures}:
\begin{equation}
\label{eq:renyi}
H_{\alpha}(Z) = \frac{1}{1-\alpha}\,\ln\!\int_{\mathbb{R}^{d}} P(z)^{\alpha}\,dz,
\qquad \alpha>0,\ \alpha\neq 1, 
\end{equation}
where the $P(z)$ represents the probability density function of the latent variable in $d$ dimensions, and $\alpha$ is the Rényi entropy of order.
In practice, we usually set $\alpha=2$, which applies a simple, differentiable estimator using pairwise Gaussian kernels~\cite{chen2019mee,2025meewireless}. 
Therefore, the second order of Rényi entropy is calculated as, $H_2(X) =-\ln\int P_X(x)^2dx$.

However, since the true probability density $P(z)$ of latent embeddings is unknown, we approximate it non-parametrically through kernel density estimation (KDE)~\cite{2021KDEsurvey}. 
KDE places a small “bump” (kernel) at each latent variable $z_i$ and averages them, avoiding any assumption that the data follow a specific distribution.
Among possible kernels, we choose the Gaussian kernel because it is smoothly differentiable, has closed-form products/convolutions (useful for our Rényi-2 entropy), and yields accurate, stable density estimates in practice. 
Given latent samples $\{z_i\}_{i=1}^{N}\subseteq \mathbb{R}^d$, the density function approximated by KDE is
\begin{equation}
    \hat{P}(z) = \frac{1}{N}\sum_{i=1}^N \mathcal{K}_{\sigma}(z-z_i),
\end{equation}
where $\mathcal{K}_{\sigma}$ is one-dimensional Gaussian kernel with bandwidth (kernel size) $\sigma$, $K_{\sigma}(z - z_i)
= \frac{1}{\sqrt{2\pi}\,\sigma}\,
\exp\!\left(-\frac{(z - z_i)^2}{2\sigma^{2}}\right).$ 

Next, the approximated density function $\hat{P}(z)$ is placed inside the Rényi entropy function to obtain the final MLE loss.  
Specifically, using the Gaussian convolution identity $K_\sigma * K_\sigma = K_{\sqrt{2}\sigma}$ for second-order, the integral reduces to our final MLE loss  (full derivation in the supplementary file) :
\begin{equation}
\label{eq:mle}
    \Lmle (\sigma) = -\log\left[\frac{1}{N^2}\sum_{i=1}^N\sum_{j=1}^N \mathcal{K}_{\sqrt{2}\sigma}(z_i-z_j)\right].
\end{equation}

This derivation shows that the entropy can be expressed directly through pairwise similarities between latent embeddings $z$, with $\sigma$ controlling the sensitivity of the kernel. 
The resulting MLE loss is thus a differentiable objective that penalizes dispersion in the latent space, encouraging embeddings to concentrate and suppressing the influence of outliers.

Finally, to integrate the MLE loss into the learning process, we define the overall training objective as the
weighted sum of the $\Lrec$ and the entropy regularization term $\Lmle$:
\begin{equation}
\mathcal{L} = \mathcal{L}_{\text{MSE}} + \lambda \, \mathcal{L}_{\text{MLE}}(\sigma),
\label{eq:dual_loss}
\end{equation}
where the $\lambda$ sets the entropy term’s influence and the $\sigma$ sets how sharply pairwise latent similarities decay. 
By jointly minimizing these two
losses, the model learns to accurately reconstruct the dominant normal patterns (driven by $\Lrec$), while simultaneously collapsing abnormal embeddings into the normal cluster (driven by the
$\Lmle$). 
Therefore, normal videos are reconstructed well, whereas abnormal videos are poorly reconstructed for detection.

\subsection{MLE loss Interpretability through t-SNE Visualization}
To demonstrate how the MLE loss $\Lmle$ improves normal reconstruction and degrades anomalies reconstruction, we apply t-Distributed Stochastic Neighbor Embedding (t-SNE)~\cite{yang2025improved} to map the latent space (e.g., 32 dimensions) into a 2D space while preserving local relationships.

\begin{figure}[htbp]
    \centering
    \includegraphics[width=\linewidth]{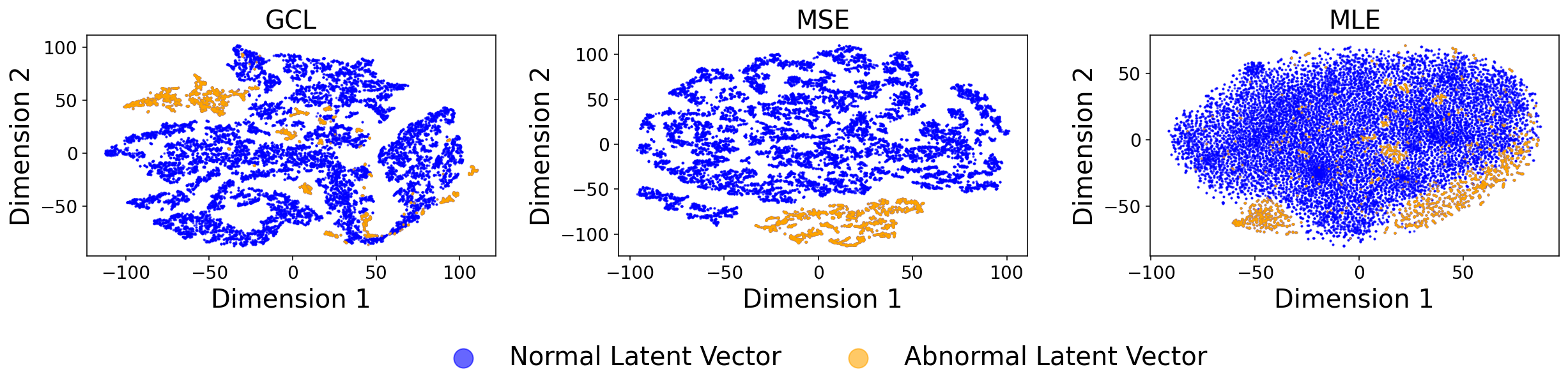}
        \vspace{-7mm}
    \caption{t-SNE visualization of latent embeddings across different methods (blue = normal, orange = abnormal). 
    Baselines, only MSE and GCL~\cite{Zaheer_GCL}, keep the normal and abnormal embeddings separate for better reconstruction.
    In contrast, our proposed MLE loss regularizes the latent distribution by collapsing the abnormal distribution into the dominant normal cluster.}
    \label{fig:overall_tsne}

\end{figure}

As shown in Fig.~\ref{fig:overall_tsne}, the embeddings of abnormal frames tend to constitute separate clusters in latent space for baselines (e.g.,  GCL~\cite{Zaheer_GCL} or autoencoder with only reconstruction loss)
This is expected: abnormal and normal videos differ inherently, so accurate reconstruction (mapping from latent to image) naturally places them in different latent regions.
Adding the MLE loss makes the latent distribution more compact (Fig.~\ref{fig:overall_tsne}: the variance along t-SNE axes decreases relative to MSE only).
As a result, entropy minimization pulls the previously separated abnormal embeddings toward the dominant normal cluster, reducing uncertainty.
Anomalies are reconstructed \textit{as if they were normal}, yielding poor reconstructions.
Although the MLE loss also slightly shifts normal clusters, its impact on normal reconstructions is minimal compared with its effect on anomalies.
Consequently, anomalies incur higher reconstruction errors, enabling a clear reconstruction error threshold for detection.

\subsection{Anomaly Detection Indicator}
After the dual-loss training process, we set a simple yet effective threshold on the reconstruction error.
The reconstruction quality is measured by the PCC between each
frame and its reconstruction (high PCC means low reconstruction error).
Since our dual loss makes the autoencoder reconstruct the anomaly poorly, anomalies lie in the \emph{lower tail} of the PCC distribution. 
Given the $\{ \text{pcc}_i \}_{i=1}^T$ across all frames in time horizon $T$, the global mean $\mu$ and standard deviation $\sigma$ will be calculated as
\begin{equation}
\mu = \frac{1}{T}\sum_{i=1}^T \text{pcc}_i, 
\qquad
\sigma = \sqrt{\frac{1}{T}\sum_{i=1}^T (\text{pcc}_i - \mu)^2},
\end{equation}

We apply a lower-tail threshold to detect anomalies:
\begin{equation}
\tau=\mu-\kappa\sigma,\qquad \text{declare $x_i$ anomaly if }~\text{pcc}_i<\tau.
\end{equation}
We set $\kappa=0.5$ for all scenarios because it consistently balances sensitivity to anomaly with a low false-alarm rate.
When PCCs are tightly concentrated (small $\sigma$), the threshold becomes stricter; when scores are more spread out (large $\sigma$), the threshold relaxes.
This variance-aware rule avoids manual threshold tuning while still reliably separating normal and abnormal frames.

\section{Experiments}

\subsection{Experimental Design and Implementation}

\textbf{Datasets.} We evaluated our approach on three datasets, covering both synthetic and real-world scenarios under different anomaly ratios. 
They provide diverse challenges: the Donkeycar dataset captures real driving with sensor noise, the Corridor dataset represents real surveillance footage with the protest abnormal events, and the UBnormal dataset introduces controlled synthetic anomalies. Further details are provided below.

\textit{Donkeycar (real-world, self-collected).}
This is a real-world dataset collected with a camera on the Donkeycar self-driving platform~\cite{2021donkeycar}.
Normal frames correspond to clean driving views, while anomalies are defined as raindrops obscuring the monitor (See Fig.~\ref{fig:overall_pcc}, bottom left).
It contains 18,000 frames (6,400 pixels per frame) with an anomaly ratio of 12.5\%. 
The moving background makes it difficult for the autoencoder to detect anomalies.

\textit{Corridor (real-world, public).}
A fixed-view surveillance dataset capturing activities in a corridor environment~\cite{rodrigues2020multi}. 
We evaluate the protest anomaly type, yielding 1,200 frames (187,200 pixels each) and an anomaly ratio of 20\% (Fig.~\ref{fig:overall_pcc}, bottom middle). The main challenges are the limited training size, which makes reconstruction less stable, and the fact that anomalies manifest as crowd events, representing realistic and complex real-world anomaly scenarios.

\textit{UBnormal (synthetic, public).}
This 3D-generated anomaly detection benchmark contains multiple types of anomalies in a virtual environment~\cite{2022UBnormal}.
We test the model in scene index 4 with the fire alarm anomaly, resulting in 902 frames (187,200 pixels each) and an anomaly ratio of 50\% (Fig.~\ref{fig:overall_pcc}, bottom right). 
The small training size and 1:1 normal–abnormal ratio make it especially challenging

\noindent
\textbf{Baseline Methods.} We compare our proposed method against three representative \textit{reconstruction-based} baseline methods:
Vanilla CAE, a standard CAE trained with MSE loss that serves as an ablation, but may also reconstruct abnormal patterns, limiting unsupervised detection performance.
GCL combines a convolutional autoencoder and a discriminator in a feedback loop~\cite{Zaheer_GCL}, where the discriminator flags poor reconstructions to guide the autoencoder;
and TMAE detects anomalies based on masked autoencoder~\cite{hu2022detecting}. FUN-AD performs fine-grained anomaly scoring and enforces local consistency in patch embeddings via mutual smoothness and a memory bank~\cite{im2025fun}. FRD-UVAD discriminates anomalies with selective cross-attention reconstruction and a disruption mechanism that suppresses abnormal features~\cite{tao2024feature}.

\noindent
\textbf{Implementation Details.} All models are trained with a CAE of embedding dimension 32 for 70 epochs using Adam ($5\times 10^{-4}$ learning rate). 
Batch sizes: 256 for Donkeycar (18k frames), 128 for Corridor (1.2k frames), and 64 for UBnormal (902 frames). 
We adopt a fully unsupervised setting in which both normal and abnormal frames are present during training; training and testing share the same unlabeled data, with labels accessed only for reporting metrics. 
After each epoch, we record PCC and AUC to monitor performance. 
To study the MLE loss, we test kernel size $\sigma$ and weight $\lambda$ over $\{0.01, 0.05, 0.1, 0.5, 1.0\}$. 
To study the robustness of our method to varying anomaly ratios, we adjust the anomaly ratio from 10\% to 60\% by randomly subsampling anomalous frames from each dataset, while keeping the total number of frames fixed.
We also evaluate generalization to unseen anomaly types by testing the trained model directly on other anomaly types.
Evaluation relies on two complementary metrics: frame-level PCC for anomaly sensitivity, AUC for overall detection accuracy.

\subsection{Experimental Results}
Our experiments answer the following questions: (A) How does the MLE loss enhance unsupervised detection? (B)  What is the impact of the hyperparameters $\sigma, \lambda$ on the detection performance? 
(C) What is the impact of varying anomaly ratios on the model's performance? (D) How well does our method generalize to different types of unseen anomalies?

\begin{table}[htbp]
\centering
\caption{AUC comparison of anomaly detectors across datasets.}
\begin{tabular}{p{2cm}lccc}
\toprule
Category & Method & Donkeycar & Corridor & UBnormal \\
\midrule
\multirow{2}{=}{OCC (Semi-supervised)} 
    & DAST~\cite{DAST_OCC}    & 0.999 & 0.996 & 0.991   \\
    & ROADMAP~\cite{wang2021robust} & 1.000 & 0.996 & 0.991   \\
    & GAN~\cite{GAN_UVAD2017} & 0.559 & 1.000 & 0.979   \\
\midrule
\multirow{5}{=}{Fully \\ Unsupervised} 
    & FUN-AD~\cite{im2025fun}                 & 0.633 & 0.478 & 0.499   \\
    & FRD-UVAD~\cite{tao2024feature}         & 0.751 & 1.000 & 0.935  \\
    & TMAE~\cite{hu2022detecting}          & 1.000 & 0.994 & 0.901   \\
    & GCL~\cite{Zaheer_GCL}           & 0.954 & 0.000 & 0.028   \\
    & Vanilla Autoencoder   & 0.691 & 0.787 & 0.066   \\
    & \cellcolor{blue!15}MLE-Guided Autoencoder (ours) & \cellcolor{blue!15} 1.000 & \cellcolor{blue!15}1.000 & \cellcolor{blue!15}1.000 \\
\bottomrule
\end{tabular}
\label{tab:auc_table}
\end{table}

\noindent
\textbf{Effect of MLE loss on full UVAD.}
With the MLE loss ($\lambda = 1, \sigma = 0.1$), our entropy-guided autoencoder consistently outperforms fully unsupervised baselines, TMAE~\cite{hu2022detecting}, GCL~\cite{Zaheer_GCL}, FUN-AD~\cite{im2025fun}, FRD-UVAD~\cite{tao2024feature}, and a vanilla autoencoder with only MSE loss, as shown in Table~\ref{tab:auc_table}.
The fully unsupervised baselines perform inconsistently. For example, TMAE~\cite{hu2022detecting} performs well on Donkeycar (AUC = 1.000) but drops noticeably on UBnormal (AUC = 0.901).
In contrast, our MLE-guided autoencoder attains near-perfect AUCs ($\approx$ 1.0) trained without labels among three datasets, matching the best OCC results (trained only on normal videos).
Furthermore, the vanilla autoencoder without MLE largely fails (e.g., AUC $\approx 0$ in UBnormal dataset).
Therefore, our proposed MLE-guided autoencoder demonstrates robust anomaly sensitivity under various conditions.

\begin{figure}[htbp]
    \centering
    \includegraphics[width=\linewidth]{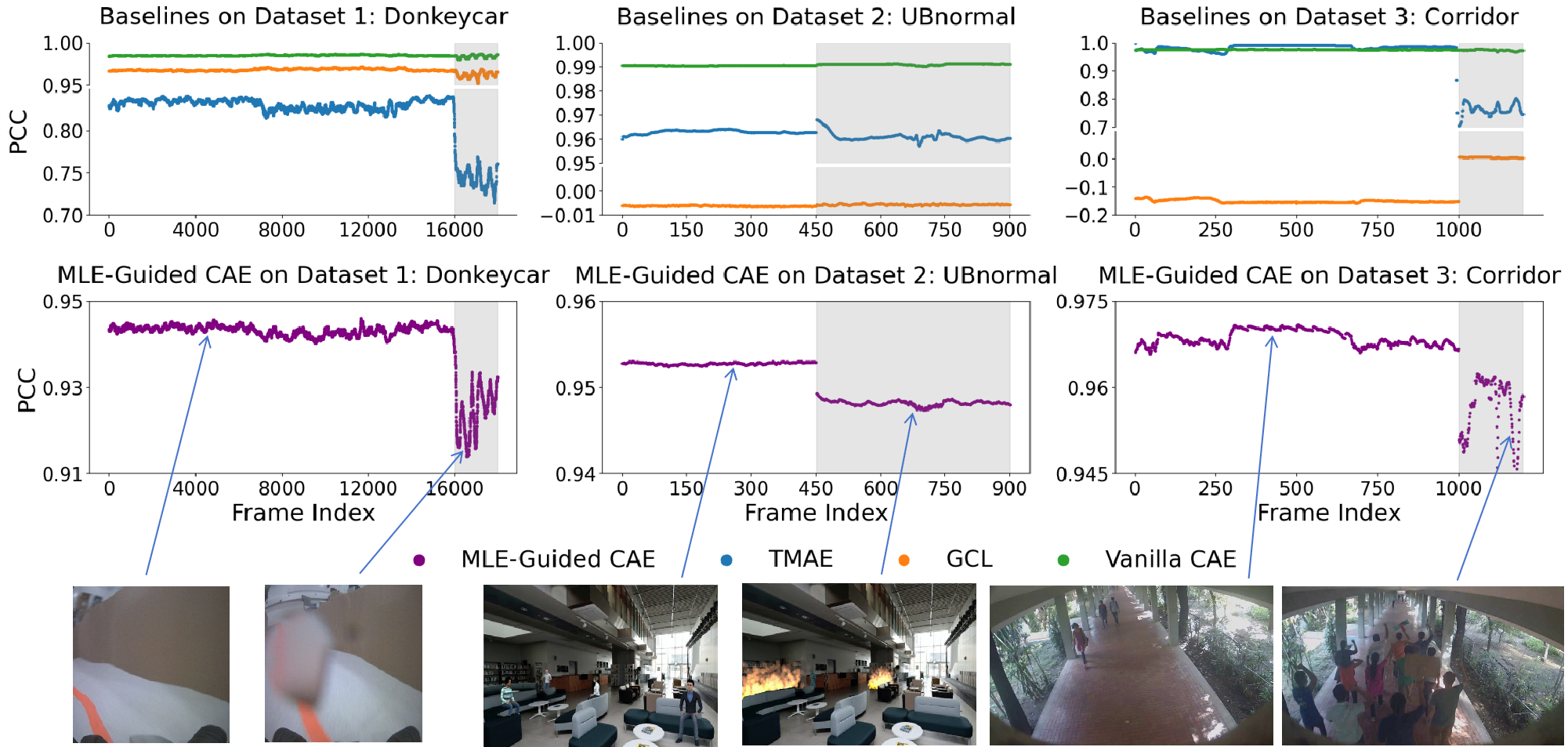}
    \caption{Unsupervised anomaly detection performance across three benchmarks. Top row: baselines (TMAE, GCL, Vanilla CAE). Bottom row: our MLE-Guided CAE. Columns: DonkeyCar, UBnormal, Corridor. Our method shows a clear normal–anomaly separation of the PCC value. }

    \label{fig:overall_pcc}
\end{figure}

To further explain the effect of MLE loss beyond Table~\ref{tab:auc_table}, we plot the per-frame reconstruction quality measured with PCC across all the datasets, as shown in Fig~\ref{fig:overall_pcc}.           
Our entropy-guided CAE maintains high PCC on normal frames (e.g., $\text{PCC} \approx 0.97$ on Corridor) and shows clear drops on abnormal frames (e.g., $\text{PCC} \approx 0.95$ on Corridor), yielding a stable reconstruction gap.
This gap is large compared to per-frame variance, so the $\mu - 0.5\sigma$ threshold can be reliable for detecting anomalies.
Furthermore, this gap happens because the MLE loss reduces latent entropy and concentrates embeddings on the dominant normal distribution.
Abnormal embeddings are pulled toward the normal distribution, which prevents the decoder from overfitting to abnormal patterns. 
Abnormal inputs are reconstructed as a normal sample, producing a larger reconstruction error and lower PCC.

In contrast, the baselines fail to separate anomalies because they do not induce a reconstruction drop.
Vanilla autoencoder, which minimizes global reconstruction error, often over-reconstructs both normal and abnormal frames, leaving only small reconstruction differences between them. 
Take the UBnormal dataset as an example in Fig~\ref{fig:overall_pcc}, PCC of normal and abnormal frames remains very high for the baselines (PCC$\approx0.96$ for TMAE; PCC$\approx0.99$ for the vanilla CAE), effectively hiding anomalies.
Meanwhile, GCL exhibits unstable behavior across datasets, sometimes even collapsing to
near-random performance (e.g., PCC $\approx 0$ in the UBnormal and Corridor dataset).

\begin{figure}[htbp]
    \centering
    \includegraphics[width=\linewidth]{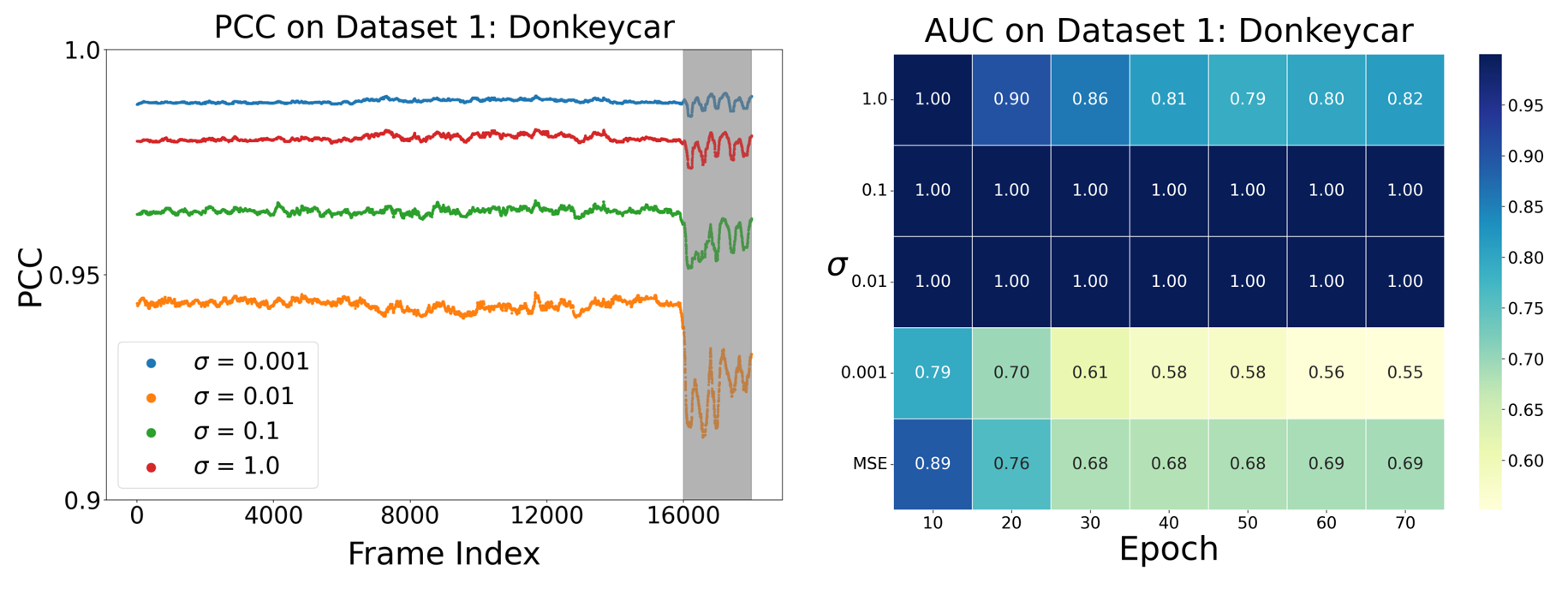}
    \caption{Effect of kernel size $\sigma$ in the MLE loss. 
    Left: PCC trajectories for different $\sigma$ values; shaded gray regions indicate ground-truth anomalies. 
    Right: AUC heatmap across epochs. 
    Mid-range $\sigma$ (0.01–0.1) yields stable reconstructions and high AUC, 
    while very small or large $\sigma$ degrades detection.}
    \label{fig:sigma_donkeycar}
\end{figure}

\noindent
\textbf{Effect of Kernel Size $\sigma$ on Detection Performance.} We evaluate detection performance across ranges of the two hyperparameters: kernel size $\sigma$ and its weight ratio $\lambda$ of MLE.
The kernel size $\sigma$ controls the smoothness of the Gaussian kernel: a very small $\sigma$ makes kernels overly sharp, whereas a very large $\sigma$ makes them nearly flat.
At either extreme $\sigma$, such as 1 or 0.001, no clear reconstruction gap exists (See blue and red PCC in Fig.~\ref{fig:sigma_donkeycar}).
Therefore, it is challenging to establish a threshold to distinguish between anomalies.
But for intermediate $\sigma$ values, a clear and easily detectable reconstruction gap emerges (See green and orange PCC in Fig.~\ref{fig:sigma_donkeycar}).
Moreover, extreme $\sigma$ values lead to unstable training and behave similarly to a vanilla autoencoder, whereas a moderate $\sigma$ yields stable training and robust separation, as shown in the AUC score heatmap in Fig.~\ref{fig:sigma_donkeycar}.

These behaviors happen because extremely small or large $\sigma$ makes the MLE loss non-informative, causing training to transfer to the reconstruction loss, and preventing abnormal embeddings from collapsing toward the normal manifold.
From a kernel density estimation (KDE) perspective, our loss induces \textit{pairwise Gaussian affinities} for embeddings,
\begin{equation}
  w_{ij} \;=\; \exp\!\Big(-\tfrac{(z_i - z_j)^2}{4\sigma^2}\Big),
  \label{eq:pairwise-weights}
\end{equation}
and the gradient on a latent embedding variable \(z_k\) is \begin{equation}
  \nabla_{z_k}\mathcal{L} \;\propto\; -\frac{1}{\sigma^2}\sum_{j}(z_k - z_j)\,w_{kj},
  \label{eq:gradient}
\end{equation}
Therefore, each latent variable is pulled toward a kernel-weighted average of its neighbors. 
When kernel size $\sigma$ is \emph{too small}, the kernels are extremely sharp and \(w_{ij}\!\approx\!0\). 
As a result, the gradient of the loss in equation \eqref{eq:gradient} becomes negligible for most \(z_k\), providing no clear direction to concentrate normals (training stalls). 
When kernel size $\sigma$ is \textit{too large}, the Gaussian kernels are nearly flat.
Since $w_{ij}\!\approx\!1$, the loss cannot tell ``near'' from ``far.'' 
Each latent embedding $z_k$ is uniformly attracted toward all the neighborhood, not toward high-density regions. 
That destroys the mechanism to cluster normals while repelling anomalies. 
Therefore, only a moderate \(\sigma\) preserves meaningful weights between near and far embedding pairs to produce informative gradients.

\begin{figure}[htbp]
    \centering
    \includegraphics[width=\linewidth]{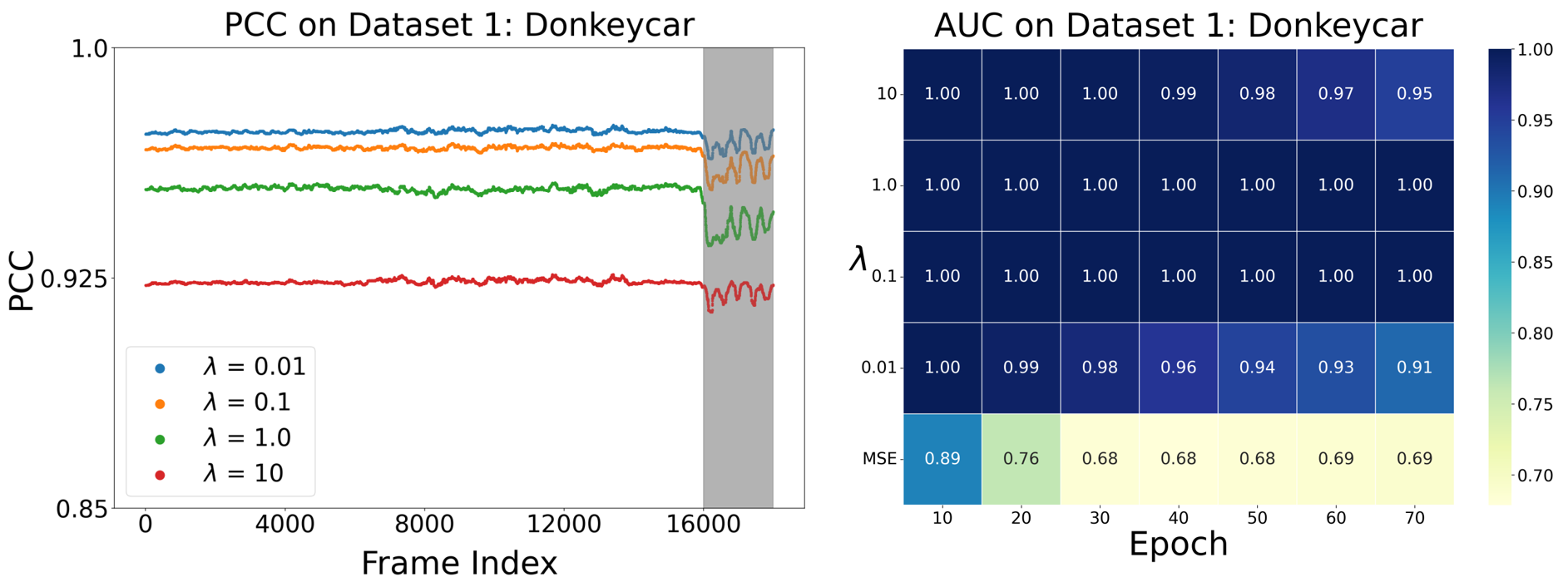}
    \caption{Effect of MLE loss weight $\lambda$ on anomaly detection. 
    Left: PCC trajectories for different $\lambda$ values; shaded gray regions indicate ground-truth anomalies. 
    Right: AUC heatmap across epochs. 
    Moderate $\lambda$ (0.1–1.0) achieves the best trade-off between stable reconstruction and clear anomaly sensitivity.}
    \label{fig:lambda_donkeycar}
\end{figure}

\noindent
\textbf{Effect of Loss Weight $\lambda$ on Detection Performance}. The weight \(\lambda\) scales the MLE loss in the total loss
$\mathcal{L}=\mathcal{L}_{\text{MSE}}+\lambda\,\mathcal{L}_{\text{MLE}}$.
When $\lambda$ is too small (e.g., $0.01$), the entropy regularization is largely ignored, making training nearly MSE-only and weakening anomaly sensitivity; this leads to reduced PCC drops on abnormal frames and lower AUC, shown in Fig.~\ref{fig:lambda_donkeycar}. 
Conversely, when $\lambda$ is too large (e.g., $10$), the entropy term dominates, over-regularizing the embeddings, which degrades reconstruction quality for both normal and abnormal frames, causing a drop in AUC. 
In contrast, mid-range values of $\lambda$ (e.g., $0.1$–$1.0$) provide the best balance: reconstructions of normal frames remain stable, while abnormal embeddings are sufficiently collapsed to yield clear PCC separation and consistently high AUC across epochs.
In short, $\lambda$ values from 0.001 to 1 and $\sigma$ values from 0.01 to 0.1 consistently achieve near-optimal AUC across all epochs. In practice, we suggest using any combination within these ranges for reliable anomaly detection.
Similar sensitivity analysis of  $\lambda$ and $\sigma$ for other datasets is provided in the supplementary material.

\begin{figure}[htbp]
    \centering
    \includegraphics[width=\linewidth]{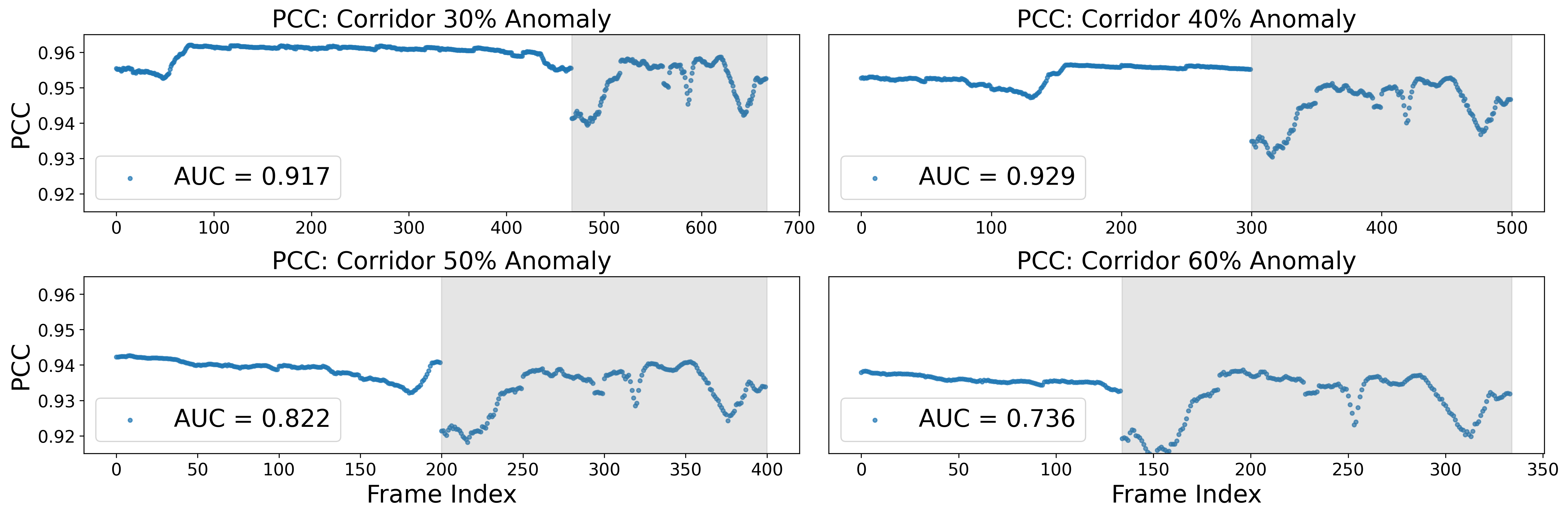}
    \caption{Impact of anomaly ratios on detection performance for the Corridor dataset. The subplots show the PCC trajectories at different anomaly ratios, with the gray regions indicating anomalies. 
    }
    \label{fig:corridor_anomaly_ratio}
\end{figure}

\noindent
\textbf{Effect of the Anomaly Ratio on Detection Performance}. 
Furthermore, we evaluated our model's performance across anomaly ratios ranging from 10\% to 60\% on all three datasets. When the anomaly ratio is below 40\%, the model maintains strong performance with AUC scores around 0.9. However, once the ratio exceeds 50\%, the AUC drops sharply to approximately 0.7 (see Fig.~\ref{fig:corridor_anomaly_ratio}).
This decline occurs because a larger anomaly ratio shifts the dominant latent-space cluster from normal to abnormal. 
Since MLE reduces overall entropy by pulling embeddings toward the main concentration, the normal cluster is dragged toward the abnormal one. 
As a result, normal frames are reconstructed poorly, and the reconstruction gap between normal and abnormal frames becomes difficult to observe. 
Nonetheless, in real-world camera systems, anomaly ratios above 50\% are rare, and we expect our model to be consistently effective with anomaly ratios below 40\%.
Additional anomaly ratio results for other datasets are provided in the supplementary material.

\begin{figure}[htbp]
    \centering
    \includegraphics[width=\linewidth]{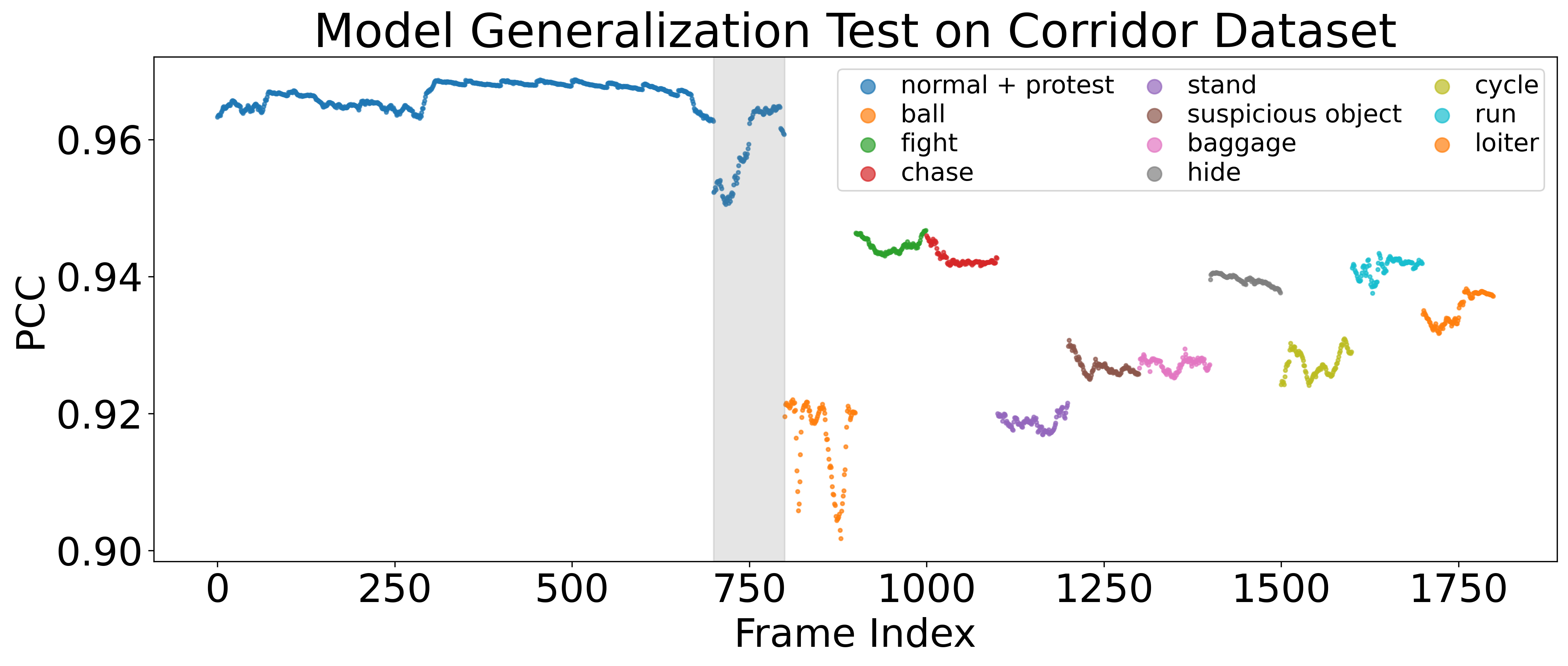}
    \caption{\looseness=-1
    Test of generalization ability across 10 different anomalies on the Corridor dataset. The model is trained on normal data with only protest abnormal data (gray area: protest) and is tested on the other 10 anomalies.
        }
    \label{fig:protest_train_transfer_testing}
\end{figure}

\noindent
\textbf{Generalization Capability Across Different Types of Anomalies.}
We evaluate the generalization ability by training our model on an unlabeled video containing normal scenes and a single anomaly type --- and testing it on the \textit{other, unseen anomalies}.
These unseen anomalies show a clear PCC drop compared to normal frames, indicating that they can be effectively detected (see Fig.~\ref{fig:protest_train_transfer_testing}). 
That is because the MLE loss encourages the model to learn the compact normal manifold by emphasizing the normal data distribution and preventing overfitting to the known anomaly during training. 
During inference, other unseen anomalies still deviate from this learned manifold and therefore cannot be reconstructed well, leading to a significantly lower PCC.
Since normal frames maintain high PCC, the resulting gap enables reliable anomaly detection.
Therefore, the MLE loss also supports strong generalization to anomaly types not observed during training.

\section{Conclusion}
\looseness=-1
This paper introduced a single-scene, fully unsupervised VAD setting and an entropy-guided autoencoder (MLE-UVAD) that adds minimal latent entropy to the reconstruction loss. This dual loss collapses sparse anomalous embeddings toward the dominant normal cluster, yielding a stable reconstruction-error gap that enables reliable, high-AUC detection in three datasets.
We further analyze the effects of key hyperparameters, different anomaly ratios, and the model’s ability to generalize to unseen anomaly types.
Our future work will deploy MLE-UVAD as a privacy-preserving data-cleaning method that outputs high-confidence normal/abnormal masks for supervised training without sharing the raw video. Also, we will extend this approach to multi-camera/multi-scene settings through domain adaptation and camera-aware priors while incorporating temporal modeling to curb false alarms from transient events.

%
%


\bibliographystyle{splncs04}
\bibliography{main}

@inproceedings{tian2021weakly,
  title={Weakly-supervised video anomaly detection with robust temporal feature magnitude learning},
  author={Tian, Yu and Pang, Guansong and Chen, Yuanhong and Singh, Rajvinder and Verjans, Johan W and Carneiro, Gustavo},
  booktitle={Proceedings of the IEEE/CVF international conference on computer vision},
  pages={4975--4986},
  year={2021}
}

@article{VADliuSurvey,
author = {Liu, Jing and Liu, Yang and Lin, Jieyu and Li, Jielin and Cao, Liang and Sun, Peng and Hu, Bo and Song, Liang and Boukerche, Azzedine and Leung, Victor C.M.},
title = {Networking Systems for Video Anomaly Detection: A Tutorial and Survey},
year = {2025},
issue_date = {October 2025},
publisher = {Association for Computing Machinery},
address = {New York, NY, USA},
volume = {57},
number = {10},
issn = {0360-0300},
url = {https://doi.org/10.1145/3729222},
doi = {10.1145/3729222},
journal = {ACM Comput. Surv.},
month = may,
articleno = {270},
numpages = {37}
}

@inproceedings{pang2020self,
  title={Self-trained deep ordinal regression for end-to-end video anomaly detection},
  author={Pang, Guansong and Yan, Cheng and Shen, Chunhua and Hengel, Anton van den and Bai, Xiao},
  booktitle={Proceedings of the IEEE/CVF conference on computer vision and pattern recognition},
  pages={12173--12182},
  year={2020}
}

@inproceedings{zong2018deep,
  title={Deep autoencoding gaussian mixture model for unsupervised anomaly detection},
  author={Zong, Bo and Song, Qi and Min, Martin Renqiang and Cheng, Wei and Lumezanu, Cristian and Cho, Daeki and Chen, Haifeng},
  booktitle={International conference on learning representations},
  year={2018}
}

@article{wang2021robust,
  title={Robust unsupervised video anomaly detection by multipath frame prediction},
  author={Wang, Xuanzhao and Che, Zhengping and Jiang, Bo and Xiao, Ning and Yang, Ke and Tang, Jian and Ye, Jieping and Wang, Jingyu and Qi, Qi},
  journal={IEEE transactions on neural networks and learning systems},
  volume={33},
  number={6},
  pages={2301--2312},
  year={2021},
  publisher={IEEE}
}

@ARTICLE{singlescenesurvay,
  author={Ramachandra, Bharathkumar and Jones, Michael J. and Vatsavai, Ranga Raju},
  journal={IEEE Transactions on Pattern Analysis and Machine Intelligence}, 
  title={A Survey of Single-Scene Video Anomaly Detection}, 
  year={2022},
  volume={44},
  number={5},
  pages={2293-2312},
  doi={10.1109/TPAMI.2020.3040591}}

@InProceedings{Zaheer_GCL,
    author    = {Zaheer, M. Zaigham and Mahmood, Arif and Khan, M. Haris and Segu, Mattia and Yu, Fisher and Lee, Seung-Ik},
    title     = {Generative Cooperative Learning for Unsupervised Video Anomaly Detection},
    booktitle = {Proceedings of the IEEE/CVF Conference on Computer Vision and Pattern Recognition (CVPR)},
    month     = {June},
    year      = {2022},
    pages     = {14744-14754}
}

@InProceedings{knowledgeinfuse_2025_CVPR,
    author    = {Dawoud, Khaled and Zaheer, Zaigham and Khan, Mustaqeem and Nandakumar, Karthik and Elsaddik, Abdulmotaleb and Khan, Muhammad Haris},
    title     = {FusedVision: A Knowledge-Infusing Approach for Practical Anomaly Detection in Real-world Surveillance Videos},
    booktitle = {Proceedings of the IEEE/CVF Conference on Computer Vision and Pattern Recognition (CVPR) Workshops},
    month     = {June},
    year      = {2025},
    pages     = {4045-4055}
}

@InProceedings{CLAP_2024_CVPR,
    author    = {Al-lahham, Anas and Zaheer, Muhammad Zaigham and Tastan, Nurbek and Nandakumar, Karthik},
    title     = {Collaborative Learning of Anomalies with Privacy (CLAP) for Unsupervised Video Anomaly Detection: A New Baseline},
    booktitle = {Proceedings of the IEEE/CVF Conference on Computer Vision and Pattern Recognition (CVPR)},
    month     = {June},
    year      = {2024},
    pages     = {12416-12425}
}

@INPROCEEDINGS{multiscene-AD,
  author={Zhang, Menghao and Wang, Jingyu and Qi, Qi and Sun, Haifeng and Zhuang, Zirui and Ren, Pengfei and Ma, Ruilong and Liao, Jianxin},
  booktitle={2024 IEEE/CVF Conference on Computer Vision and Pattern Recognition (CVPR)}, 
  title={Multi-Scale Video Anomaly Detection by Multi-Grained Spatio-Temporal Representation Learning}, 
  year={2024},
  volume={},
  number={},
  pages={17385-17394},
  doi={10.1109/CVPR52733.2024.01646}}

@inproceedings{sultani2018real,
  title={Real-world anomaly detection in surveillance videos},
  author={Sultani, Waqas and Chen, Chen and Shah, Mubarak},
  booktitle={Proceedings of the IEEE conference on computer vision and pattern recognition},
  pages={6479--6488},
  year={2018}
}

@inproceedings{zhong2019graph,
  title={Graph convolutional label noise cleaner: Train a plug-and-play action classifier for anomaly detection},
  author={Zhong, Jia-Xing and Li, Nannan and Kong, Weijie and Liu, Shan and Li, Thomas H and Li, Ge},
  booktitle={Proceedings of the IEEE/CVF conference on computer vision and pattern recognition},
  pages={1237--1246},
  year={2019}
}

@article{wan2021weakly,
  title={Weakly supervised video anomaly detection via center-guided discriminative learning},
  author={Wan, Boyang and Fang, Yuming and Xia, Xue and Mei, Jiajie},
  journal={arXiv preprint arXiv:2104.07268},
  year={2021}
}

@inproceedings{hu2022detecting,
  title={Detecting anomalous events from unlabeled videos via temporal masked auto-encoding},
  author={Hu, Jingtao and Yu, Guang and Wang, Siqi and Zhu, En and Cai, Zhiping and Zhu, Xinzhong},
  booktitle={2022 IEEE International Conference on Multimedia and Expo (ICME)},
  pages={1--6},
  year={2022},
  organization={IEEE}
}

@article{yang2025improved,
  title={Improved PCA Reconstruction-Based Unsupervised Anomaly Detection in Uncontrolled Structural Health Monitoring With Correntropy},
  author={Yang, Kang and Lin, Zhenhan and Yang, Zekun and Tian, Zhihui and Ma, Jie and Pr{\'\i}ncipe, Jos{\'e} C and Harley, Joel B},
  journal={IEEE Transactions on Industrial Informatics},
  year={2025},
  publisher={IEEE}
}

@inproceedings{reconstructionAD-oldest,
author = {Zhou, Chong and Paffenroth, Randy C.},
title = {Anomaly Detection with Robust Deep Autoencoders},
year = {2017},
isbn = {9781450348874},
publisher = {Association for Computing Machinery},
address = {New York, NY, USA},
url = {https://doi.org/10.1145/3097983.3098052},
doi = {10.1145/3097983.3098052},
booktitle = {Proceedings of the 23rd ACM SIGKDD International Conference on Knowledge Discovery and Data Mining},
pages = {665–674},
numpages = {10},
location = {Halifax, NS, Canada},
series = {KDD '17}
}

@article{DAST_OCC,
  title={DAST-Net: Dense visual attention augmented spatio-temporal network for unsupervised video anomaly detection},
  author={Kommanduri, Rangachary and Ghorai, Mrinmoy},
  journal={Neurocomputing},
  volume={579},
  pages={127444},
  year={2024},
  publisher={Elsevier}
}

@inproceedings{2024realweakly,
  title={Real-time weakly supervised video anomaly detection},
  author={Karim, Hamza and Doshi, Keval and Yilmaz, Yasin},
  booktitle={Proceedings of the IEEE/CVF winter conference on applications of computer vision},
  pages={6848--6856},
  year={2024}
}

@inproceedings{2024VADreview,
 author = {Zhu, Liyun and Wang, Lei and Raj, Arjun and Gedeon, Tom and Chen, Chen},
 booktitle = {Advances in Neural Information Processing Systems},
 editor = {A. Globerson and L. Mackey and D. Belgrave and A. Fan and U. Paquet and J. Tomczak and C. Zhang},
 pages = {89943--89977},
 publisher = {Curran Associates, Inc.},
 title = {Advancing Video Anomaly Detection: A Concise Review and a New Dataset},
 url = {https://proceedings.neurips.cc/paper_files/paper/2024/file/a3c5af1f56fc73eef1ba0f442739f5ca-Paper-Datasets_and_Benchmarks_Track.pdf},
 volume = {37},
 year = {2024}
}

@article{gan2predict2024,
title = {Dual GroupGAN: An unsupervised four-competitor (2V2) approach for video anomaly detection},
journal = {Pattern Recognition},
volume = {153},
pages = {110500},
year = {2024},
issn = {0031-3203},
doi = {https://doi.org/10.1016/j.patcog.2024.110500},
url = {https://www.sciencedirect.com/science/article/pii/S0031320324002516},
author = {Zhe Sun and Panpan Wang and Wang Zheng and Meng Zhang}
}

@inproceedings{GAN_UVAD2017,
  title={Abnormal event detection in videos using generative adversarial nets},
  author={Ravanbakhsh, Mahdyar and Nabi, Moin and Sangineto, Enver and Marcenaro, Lucio and Regazzoni, Carlo and Sebe, Nicu},
  booktitle={2017 IEEE international conference on image processing (ICIP)},
  pages={1577--1581},
  year={2017},
  organization={IEEE}
}

@inproceedings{2022UBnormal,
  title={Ubnormal: New benchmark for supervised open-set video anomaly detection},
  author={Acsintoae, Andra and Florescu, Andrei and Georgescu, Mariana-Iuliana and Mare, Tudor and Sumedrea, Paul and Ionescu, Radu Tudor and Khan, Fahad Shahbaz and Shah, Mubarak},
  booktitle={Proceedings of the IEEE/CVF conference on computer vision and pattern recognition},
  pages={20143--20153},
  year={2022}
}

@inproceedings{rodrigues2020multi,
  title={Multi-timescale trajectory prediction for abnormal human activity detection},
  author={Rodrigues, Royston and Bhargava, Neha and Velmurugan, Rajbabu and Chaudhuri, Subhasis},
  booktitle={Proceedings of the IEEE/CVF winter conference on applications of computer vision},
  pages={2626--2634},
  year={2020}
}

@inproceedings{geng2025unsupervised,
  author    = {Yuang Geng and Yang Zhou and Yuyang Zhang and Zhongzheng Ren Zhang and Kang Yang and Tyler Ruble and Giancarlo Vidal and Ivan Ruchkin},
  title     = {Unsupervised Anomaly Detection Improves Imitation Learning for Autonomous Racing},
  booktitle = {Proceedings of the {IEEE/RSJ} International Conference on Intelligent Robots and Systems ({IROS})},
  year      = {2025},
  publisher = {IEEE},
  note      = {to appear}
}

@INPROCEEDINGS{2021donkeycar,
  author={Viitala, Ari and Boney, Rinu and Zhao, Yi and Ilin, Alexander and Kannala, Juho},
  booktitle={2021 20th International Conference on Advanced Robotics (ICAR)}, 
  title={Learning to Drive (L2D) as a Low-Cost Benchmark for Real-World Reinforcement Learning}, 
  year={2021},
  volume={},
  number={},
  pages={275-281},
  doi={10.1109/ICAR53236.2021.9659342}}

@ARTICLE{2025privacyVADsurvey,
  author={Liu, Yang and Liu, Siao and Zhu, Xiaoguang and Yang, Hao and Li, Jielin and Guo, Juncen and Teng, Liangyu and Yang, Dingkang and Wang, Yan and Liu, Jing},
  journal={IEEE Transactions on Neural Networks and Learning Systems}, 
  title={Privacy-Preserving Video Anomaly Detection: A Survey}, 
  year={2025},
  volume={},
  number={},
  pages={1-22},
  doi={10.1109/TNNLS.2025.3600252}}

@article{2021KDEsurvey,
  title={Random features for kernel approximation: A survey on algorithms, theory, and beyond},
  author={Liu, Fanghui and Huang, Xiaolin and Chen, Yudong and Suykens, Johan AK},
  journal={IEEE Transactions on Pattern Analysis and Machine Intelligence},
  volume={44},
  number={10},
  pages={7128--7148},
  year={2021},
  publisher={IEEE}
}

@inproceedings{renyi1961measures,
  title={On measures of entropy and information},
  author={R{\'e}nyi, Alfr{\'e}d},
  booktitle={Proceedings of the fourth Berkeley symposium on mathematical statistics and probability, volume 1: contributions to the theory of statistics},
  volume={4},
  pages={547--562},
  year={1961},
  organization={University of California Press}
}

@article{2019conceptYU,
  title={Concept drift detection and adaptation with hierarchical hypothesis testing},
  author={Yu, Shujian and Abraham, Zubin and Wang, Heng and Shah, Mohak and Wei, Yantao and Pr{\'\i}ncipe, Jos{\'e} C},
  journal={Journal of the Franklin Institute},
  volume={356},
  number={5},
  pages={3187--3215},
  year={2019},
  publisher={Elsevier}
}

@inproceedings{2023VAEcasual,
  title={Causal recurrent variational autoencoder for medical time series generation},
  author={Li, Hongming and Yu, Shujian and Principe, Jose},
  booktitle={Proceedings of the AAAI conference on artificial intelligence},
  volume={37},
  number={7},
  pages={8562--8570},
  year={2023}
}

@article{multisceneUVA2024,
  title={Contextual information based anomaly detection for multi-scene aerial videos},
  author={Verma, Ujjwal and Pai, Manohara M M and Pai, Radhika M and others},
  journal={Scientific Reports},
  volume={15},
  number={1},
  pages={1--18},
  year={2025},
  publisher={Nature Publishing Group}
}

@inproceedings{2024entropy,
  title={Rethinking information-theoretic generalization: Loss entropy induced PAC bounds},
  author={Dong, Yuxin and Gong, Tieliang and Chen, Hong and Yu, Shujian and Li, Chen},
  booktitle={The Twelfth International Conference on Learning Representations},
  year={2024}
}

@article{chen2019mee,
  title={Minimum error entropy Kalman filter},
  author={Chen, Badong and Dang, Lujuan and Gu, Yuantao and Zheng, Nanning and Pr{\'\i}ncipe, Jos{\'e} C},
  journal={IEEE Transactions on Systems, Man, and Cybernetics: Systems},
  volume={51},
  number={9},
  pages={5819--5829},
  year={2019},
  publisher={IEEE}
}

@inproceedings{2025meewireless,
  title={An analysis of minimum error entropy loss functions in wireless communications},
  author={Pallewela, Rumeshika and Eldeeb, Eslam and Alves, Hirley},
  booktitle={2025 IEEE 101st Vehicular Technology Conference (VTC2025-Spring)},
  pages={1--6},
  year={2025},
  organization={IEEE}
}

@inproceedings{im2025fun,
  title={FUN-AD: Fully Unsupervised Learning for Anomaly Detection with Noisy Training Data},
  author={Im, Jiin and Son, Yongho and Hong, Je Hyeong},
  booktitle={2025 IEEE/CVF Winter Conference on Applications of Computer Vision (WACV)},
  pages={9447--9456},
  year={2025},
  organization={IEEE}
}

@article{tao2024feature,
  title={Feature reconstruction with disruption for unsupervised video anomaly detection},
  author={Tao, Chenchen and Wang, Chong and Lin, Sunqi and Cai, Suhang and Li, Di and Qian, Jiangbo},
  journal={IEEE Transactions on Multimedia},
  volume={26},
  pages={10160--10173},
  year={2024},
  publisher={IEEE}
}

@String(CVPR  = {IEEE Conf. Comput. Vis. Pattern Recog.})

@String(AAAI  = {AAAI})

@String(ICIP  = {IEEE Int. Conf. Image Process.})

@String(ICME  = {Int. Conf. Multimedia and Expo})

@String(PR    = {Pattern Recognition})

@String(CVPR  = {CVPR})

@String(ICIP  = {ICIP})

@String(ICME  =	{ICME})

@String(PR    = {PR})

@inproceedings{zhang2024trafficnight,
  title={TrafficNight: An aerial multimodal benchmark for nighttime vehicle surveillance},
  author={Zhang, Guoxing and Liu, Yiming and Yang, Xiaoyu and Huang, Hailong and Huang, Chao},
  booktitle={European Conference on Computer Vision},
  pages={36--48},
  year={2024},
  organization={Springer}
}

@article{2024clustering,
  title={Video anomaly detection guided by clustering learning},
  author={Qiu, Shaoming and Ye, Jingfeng and Zhao, Jiancheng and He, Lei and Liu, Liangyu and Huang, Xinchen and others},
  journal={Pattern Recognition},
  volume={153},
  pages={110550},
  year={2024},
  publisher={Elsevier}
}

\end{document}